\documentclass[10pt,twocolumn,letterpaper]{article}

\usepackage[pagenumbers]{cvpr} 

\usepackage{graphicx}
\usepackage{amsmath}
\usepackage{amssymb}
\usepackage{booktabs}
\usepackage[table]{xcolor}
\usepackage{tablefootnote}
\newcommand{\mb}[1]{\mathbf{#1}}
\newcommand{\bs}[1]{\boldsymbol{#1}}
\newcommand{\bm}[1]{\boldsymbol{\mathcal{#1}}}
\usepackage{minipage-marginpar}

\usepackage[pagebackref,breaklinks,colorlinks]{hyperref}

\usepackage[capitalize]{cleveref}
\crefname{section}{Sec.}{Secs.}
\Crefname{section}{Section}{Sections}
\Crefname{table}{Table}{Tables}
\crefname{table}{Tab.}{Tabs.}

\def\cvprPaperID{4523} 
\def\confName{CVPR}
\def\confYear{2023}

\begin{document}


\title{Mutual Information-Based Temporal Difference Learning for \\ Human Pose Estimation in Video}

\author{
    Runyang Feng\textsuperscript{\rm 1,2},
    Yixing Gao\textsuperscript{\rm 1,2}\thanks{Corresponding Author},
    Xueqing Ma\textsuperscript{\rm 1,2},
    Tze Ho Elden Tse\textsuperscript{\rm 3},
    Hyung Jin Chang\textsuperscript{\rm 3}
    \\    
    \textsuperscript{1} School of Artificial Intelligence, Jilin University,\\
    \textsuperscript{2} Engineering Research Center of Knowledge-Driven Human-Machine Intelligence, \\Ministry of Education, China,
    \textsuperscript{3}School of Computer Science, University of Birmingham
	\\
	{\tt\small \{fengry22, maxq21\}@mails.jlu.edu.cn,
	 \tt\small gaoyixing@jlu.edu.cn,}\\
	 {\tt\small txt994@student.bham.ac.uk, h.j.chang@bham.ac.uk
	}
}


\maketitle

\begin{abstract}
   Temporal modeling is crucial for multi-frame human pose estimation. 
   Most existing methods directly employ optical flow or deformable convolution to predict full-spectrum motion fields, which might incur numerous irrelevant cues, such as a nearby person or background. Without further efforts to excavate meaningful motion priors, their results are suboptimal, especially in complicated spatio-temporal interactions. 
   On the other hand, the temporal difference has the ability to encode representative motion information which can potentially be valuable for pose estimation but has not been fully exploited.
	 In this paper, we present a novel multi-frame human pose estimation framework, which employs temporal differences across frames to model dynamic contexts and engages mutual information objectively to facilitate useful motion information disentanglement. 
	To be specific, we design a multi-stage Temporal Difference Encoder that performs incremental cascaded learning conditioned on multi-stage feature difference sequences to derive informative motion representation. 
	We further propose a Representation Disentanglement module from the mutual information perspective, which can grasp discriminative task-relevant motion signals by explicitly defining useful and noisy constituents of the raw motion features and minimizing their mutual information.
	These place
	us to rank No.1 in the Crowd Pose Estimation in Complex Events Challenge on benchmark dataset HiEve, and achieve state-of-the-art performance on three  benchmarks PoseTrack2017, PoseTrack2018, and PoseTrack21.
   
	    


\end{abstract}
\begin{figure}
\begin{center}
\includegraphics[width=0.98\linewidth]{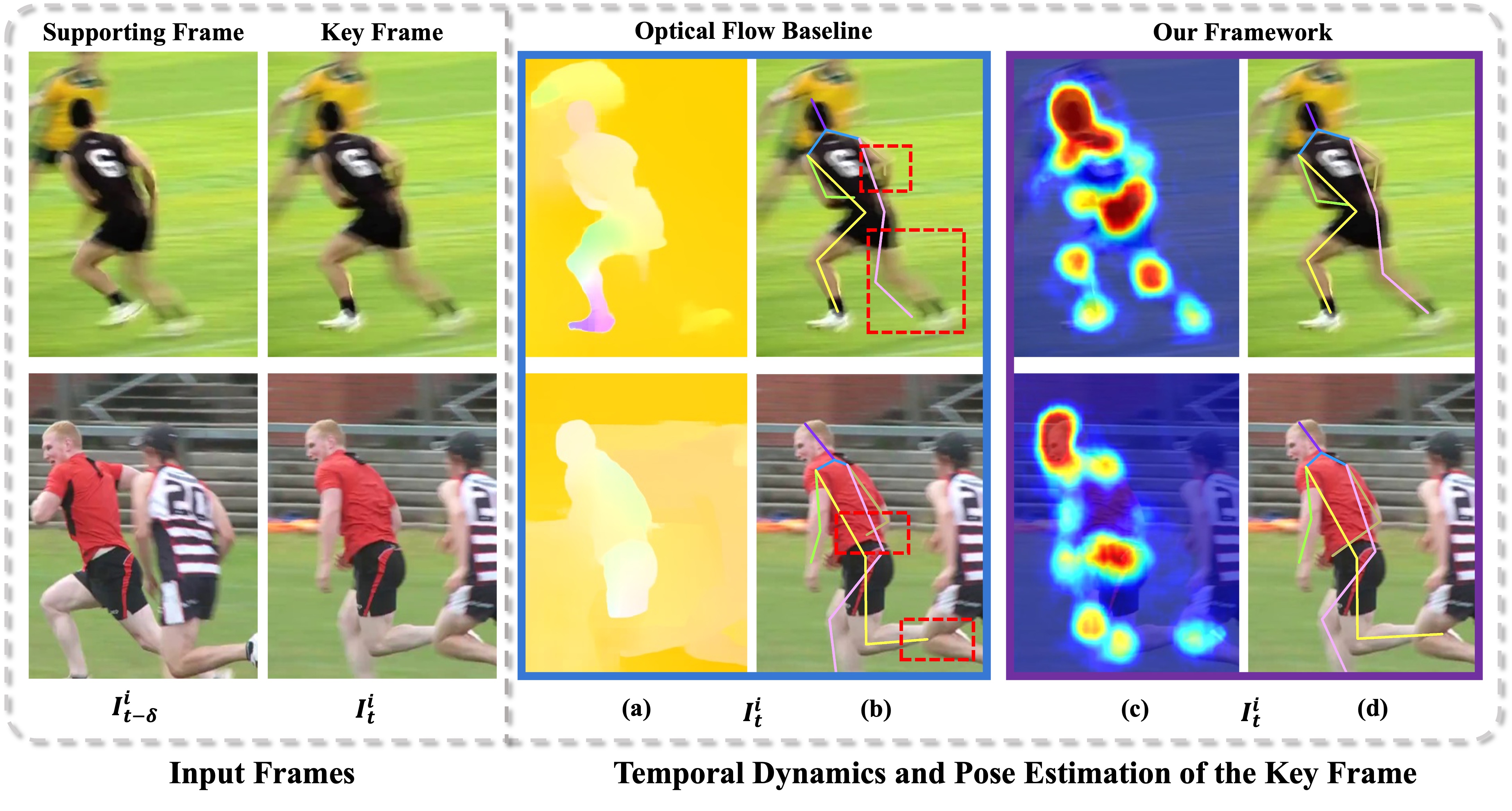}
\end{center}
\vspace{-1.5em}
\caption{
Directly leveraging optical flow can be distracted by irrelevant clues such as background and  blur 
(a), and sometimes fails in scenarios with fast motion and mutual occlusion (b). 
Our proposed framework proceeds with temporal difference encoding and useful information disentanglement to capture more tailored temporal dynamics (c),  yielding more robust pose estimations (d).
}\vspace{-1.5em}
\label{fig:motivation}
\end{figure}
\vspace{-1.em}
\section{Introduction}
\label{sec:intro}
Human pose estimation has long been a nontrivial and fundamental problem in the computer vision community. The goal is to localize anatomical keypoints (\emph{e.g.}, nose, ankle, etc.) of human bodies from images or videos. Nowadays, as more and more videos are recorded endlessly, video-based human pose estimation has been  extremely desired in enormous applications including live streaming, augmented reality, surveillance, and movement tracking \cite{isobe2022look, schmidtke2021unsupervised, liu2022temporal}.

An extensive body of literature focuses 
on estimating human poses in \emph{static images}, ranging from earlier methods employing pictorial structure models \cite{wang2008multiple, wang2013beyond, zhang2009efficient, sapp2010cascaded} to recent attempts leveraging deep convolutional neural networks \cite{Toshev_2014_CVPR, Wei_2016_CVPR,  xiao2018simple, liu2022temporal} or Vision Transformers \cite{li2021tokenpose, yuan2021hrformer, yang2021transpose}. Despite the impressive performance in still images, the extension of such methods to video-based human pose estimation 
still remains 
challenging due to the additional temporal dimension in videos \cite{wu2022motion, liu2021deep}. 
By nature, the video presents distinctive and valuable \emph{dynamic contexts} (\emph{i.e.}, the temporal evolution in the visual content) \cite{zhao2018recognize}. Therefore, being able to effectively utilize the temporal dynamics (motion information) is fundamentally important for accurate pose estimation in videos \cite{liu2022temporal}.

One line of work \cite{liu2022temporal, wang2020combining, luo2018lstm} attempts to derive a unified spatial-temporal representation through \emph{implicit} motion compensation. \cite{wang2020combining} presents a 3DHRNet which utilizes 3D convolutions to extract spatiotemporal features of a video tracklet to estimate pose sequences. \cite{liu2022temporal} adopts deformable convolutions to align multi-frame features and aggregates aligned feature maps to predict  human poses.
	On the other hand, 
	\cite{song2017thin, pfister2015flowing, zhang2018poseflow} \emph{explicitly} model motion contexts with optical flow. 
    \cite{song2017thin, pfister2015flowing} propose to compute dense optical flow between every two frames and leverage the flow features for refining pose heatmaps temporally across multiple frames. 

Upon studying the previous methods \cite{liu2021deep, liu2022temporal, pfister2015flowing, song2017thin}, we empirically observe that the pose estimation performance is boosted with the implicit or explicit imposition of motion priors. However, the movement of any visual evidence is usually attended to in these paradigms, resulting in cluttered motion features that include numerous irrelevant information (\emph{e.g.}, nearby person, background), as illustrated in Fig. \ref{fig:motivation}. Directly exploiting such vanilla motion features delivers inferior results, especially in complex scenarios of mutual occlusion and fast motion. More specifically, not all pixel movements are equally important in video-based human pose estimation \cite{zeng2022not}. For example, background variations and pixel changes caused by image quality degradation (\emph{e.g.}, blur and occlusion) are usually  useless and distracting, whereas the salient pixel movements driven by human body motions play a more important role in understanding motion patterns \cite{huang2021self}. Therefore, discovering meaningful motion dynamics is crucial to fully recovering human poses  across a video.
On the other hand, investigating temporal differences across video frames allows one to discover representative motion cues \cite{wang2021tdn, xiao2022learning, jiang2019stm}. Although it has already shown success in various video-related tasks (action recognition \cite{wang2021tdn}, video super-resolution \cite{isobe2022look}), its application on video-based human pose estimation remains under-explored.


In this paper, we present a novel framework, named \underline{\textbf{T}}emporal \underline{\textbf{D}}ifference Learning based on \underline{\textbf{M}}utual \underline{\textbf{I}}nformation (TDMI) for human pose estimation.
	Our TDMI consists of two key components: 
	  \textbf{(i)} A multi-stage Temporal Difference Encoder (TDE) is designed to model motion contexts conditioned on multi-stage feature differences among video frames. Specifically, we first compute the feature difference sequences across multiple stages by leveraging a temporal difference operator. Then, we perform incremental cascaded learning via intra- and inter-stage feature integration to derive the motion representation. 
	  \textbf{(ii)} We further introduce a Representation Disentanglement module (RDM) from the mutual information perspective, which distills the task-relevant motion features to enhance the frame representation for pose estimation. In particular, we first disentangle the useful and noisy constituents of the vanilla  motion representation by activating corresponding feature channels. Then, we theoretically analyze the statistical dependencies between the useful and the noisy motion features and arrive at an information-theoretic loss. Minimizing this mutual information objective encourages the useful motion components to be more discriminative and task-relevant. Our approach achieves significant and consistent performance improvements over current state-of-the-art methods on four benchmark datasets. Extensive ablation studies are conducted to validate the efficacy of each component in the proposed method.

The main contributions of this work can be summarized as follows: (1) We propose a novel framework that leverages temporal differences to model dynamic contexts for video-based human pose estimation.
(2) We present a disentangled representation learning strategy to grasp discriminative task-relevant motion signals via an information-theoretic objective.
(3) We demonstrate that our approach achieves new state-of-the-art results on four benchmark datasets, PoseTrack2017, PoseTrack2018, PoseTrack21, and HiEve. 


\section{Related Work}
\textbf{Image-based human pose estimation.}\quad
With the recent advances in deep learning architectures \cite{he2016deep, vaswani2017attention} as well as the availability of large-scale datasets \cite{Iqbal_2017_CVPR, Andriluka_2018_CVPR, doering2022posetrack21, lin2020human},
various deep learning methods  \cite{artacho2020unipose, cheng2020higherhrnet, sun2019deep, xiao2018simple, li2021tokenpose, yuan2021hrformer, yang2021transpose} are proposed for image-based human pose estimation. These approaches broadly fall into two paradigms: bottom-up and top-down. \emph{Bottom-up} approaches \cite{Cao_2017_CVPR, kocabas2018multiposenet, kreiss2019pifpaf} detect individual body parts and  associate them with an entire person. \cite{kreiss2019pifpaf} proposes a composite framework that employs a Part Intensity Field to localize human body parts and uses a Part Association Field to associate the detected body parts with each other. Conversely, \emph{top-down} approaches \cite{xiao2018simple, Wei_2016_CVPR, sun2019deep, fang2017rmpe, li2021tokenpose} detect bounding boxes of persons first and  predict human poses within each bounding box region. 
\cite{sun2019deep} presents a high-resolution convolutional architecture that preserves high-resolution features in all stages, demonstrating superior performance for human pose estimation.

\textbf{Video-based human pose estimation.}\quad
Existing image-based methods could not generalize well to video streams 
since they inherently have difficulties in capturing temporal dynamics across frames. 
A direct approach would be to leverage optical flow to impose motion priors \cite{song2017thin, pfister2015flowing}.
These approaches typically compute dense optical flow among frames and leverage such motion cues to refine the predicted pose heatmaps. 
However, the optical flow estimation is computationally intensive and tends to be vulnerable when encountering severe image quality degradation. Another approach \cite{liu2022temporal, bertasius2019learning, liu2021deep, wang2020combining} considers implicit motion compensation using deformable convolutions or 3DCNNs. \cite{bertasius2019learning, liu2021deep} propose to model multi-granularity joint movements based on heatmap residuals and perform pose resampling or pose warping through deformable convolutions. 
As the above cases generally consider motion details from all pixel locations, their resulting representations are suboptimal for accurate pose estimation. 




\textbf{Temporal difference modeling.}\quad
Temporal difference operations, \emph{i.e.}, RGB Difference (image-level) \cite{wang2016temporal, zhao2018recognize, ng2018temporal, wang2021tdn} and Feature Difference (feature-level) \cite{liu2020teinet, jiang2019stm, li2020tea}, are typically exploited for motion extraction,  showing outstanding performance with high efficiency for many video-related tasks such as action recognition \cite{wang2021tdn, li2020tea} and video super-resolution \cite{isobe2022look}. \cite{zhao2018recognize, ng2018temporal, wang2021tdn} leverage RGB difference as an efficient alternative modality to optical flow to represent motions. \cite{isobe2022look} proposes to explicitly model temporal differences in both LR and HR space. However, the additional RGB difference branch usually replicates the feature-extraction backbone, which increases the model complexity. 
On the other hand, \cite{liu2020teinet, jiang2019stm, li2020tea} employ a feature difference operation for network design which our work falls within this scope more closely.
In contrast to previous methods that simply compute feature differences, we seek to disentangle discriminative task-relevant temporal difference representations for pose estimation.
\vspace{-0.4em}
\section{Our Approach}
\textbf{Preliminaries.}\quad 
Our work follows the top-down paradigm, starting with an object detector to obtain the bounding boxes for individual  persons in a video frame $I_t$. Then, each bounding box is enlarged by $25\%$ to crop the same individual in a consecutive frame sequence $\boldsymbol{\mathcal{X}}_t = \left \langle I_{t-\delta},...,I_t,...,I_{t+\delta} \right \rangle$ with $\delta$ being a predefined temporal span. In this way, we attain the cropped video clip $\boldsymbol{\mathcal{X}}^i_t = \left \langle I_{t-\delta}^i,...,I_t^i,...,I_{t+\delta}^i \right \rangle$ for person $i$.

\textbf{Problem formulation.}\quad 
Given a cropped video segment $\boldsymbol{\mathcal{X}}^i_t$ centered on the key frame $I_t^i$, we are interested in estimating the pose in $I_t^i$. Our goal is to better leverage frame sequences through principled temporal difference learning and useful information disentanglement, thereby addressing the common shortcoming of existing methods in failing to adequately mine motion dynamics.

\textbf{Method overview.}\quad
The overall pipeline of the proposed TDMI is outlined in Fig. \ref{fig:pipeline}. Our framework  consists of two key components: a multi-stage Temporal Difference Encoder (TDE) (Sec. \ref{sec. smtde}) and a Representation Disentanglement module (RDM) (Sec. \ref{sec. drl}). Specifically, we first extract visual features of the input sequence and feed them to TDE, which computes feature differences and performs information integration to obtain the motion feature $\mb{M}_t^i$. Then, RDM takes motion feature $\mb{M}_t^i$ as input and excavates its useful constituents to yield $\mb{M}_{t,u}^{i}$. Finally, both the motion feature $\mb{M}_{t,u}^{i}$ and the visual feature of the key frame are aggregated to produce the enhanced representation $ \tilde {\mb{F}}_t^i$.  $ \tilde {\mb{F}}_t^i$ is handed to a detection head which outputs the pose estimation $\mb{H}_t^i$. In the following, we explain the two key components in depth.

\vspace{-0.4em}
\subsection{Multi-Stage Temporal Difference Encoder} \label{sec. smtde}
\vspace{-0.4em}
As multi-stage feature integration enables the network to retain diverse semantic information from fine to coarse scale \cite{jiang2020pay, pang2019towards, zeng2022not}, we propose to simultaneously aggregate shallow feature differences (early stages) compressing detailed motion cues and deep feature differences (late stages) encoding global semantic movements to derive informative and fine-grained motion representations. 
A naive approach to fuse features in multiple stages is to feed them into a convolutional network \cite{ke2018multi, chen2018cascaded}. However, this simple fusion solution suffers from two drawbacks: (i) redundant features might be over-emphasized, and (ii) fine-grained cues of each stage cannot be fully reserved. Motivated by these observations and insights, we present a multi-stage temporal difference encoder (TDE) with an incremental cascaded learning architecture, addressing the above issues through two designs: a \emph{spatial modulation} mechanism to adaptively focus on important information at each stage, and a \emph{progressive accumulation} mechanism to preserve fine-grained contexts across all stages.

Specifically, given an image sequence $\boldsymbol{\mathcal{X}}^i_t = \left \langle I_{t-\delta}^i,...,I_t^i,...,I_{t+\delta}^i \right \rangle$, our  proposed TDE first constructs multi-stage feature difference sequences and performs both intra- and inter-stage feature fusion to yield the encoded motion representation $\mb{M}_t^i$. For simplicity, we take $\delta = 1$ in the following.

\begin{figure*}
\begin{center}
\includegraphics[width=0.96\linewidth]{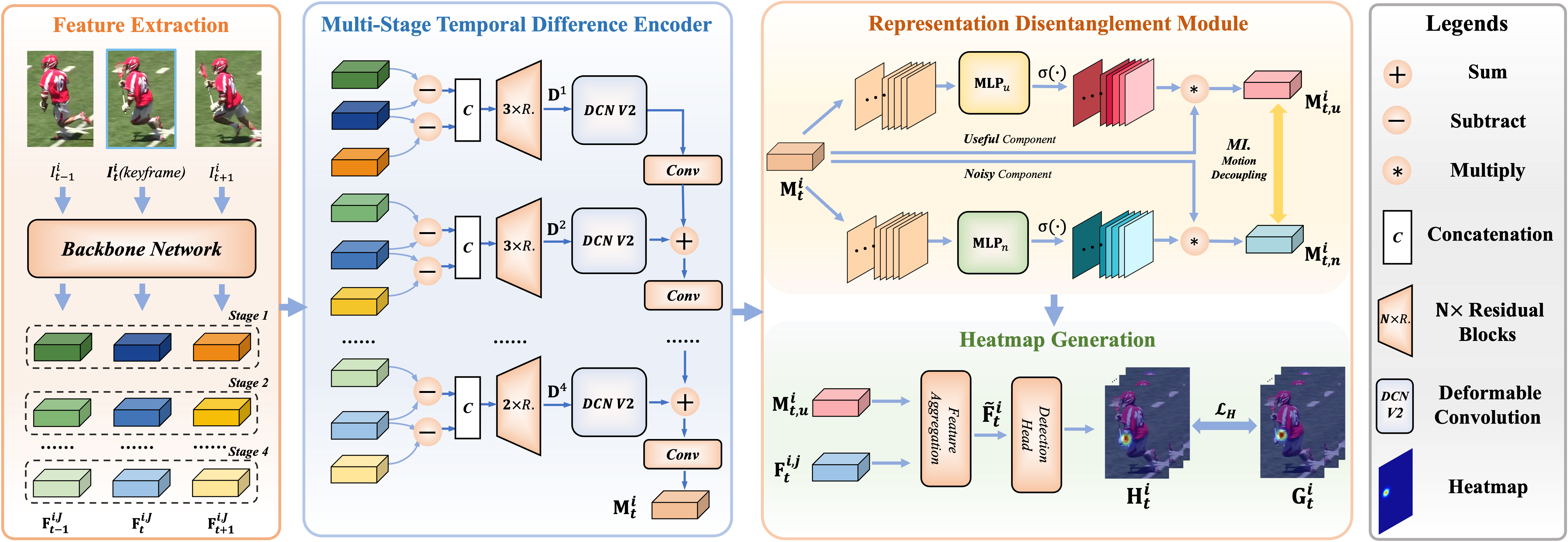}
\end{center}
\vspace{-1em}
\caption{{Overall pipeline of the proposed framework. The goal is to detect the human pose of the key frame $I_t^i$. Given an input sequence $ \left \langle I_{t-1}^i,I_t^i,I_{t+1}^i \right \rangle$, we first extract their visual features $\{ \mb{F}_{t-1}^{i,J}, \mb{F}_{t}^{i,J}, \mb{F}_{t+1}^{i,J} \}$. Our multi-stage Temporal Difference Encoder  takes these features as input and outputs the motion feature $\mb{M}_t^i$. Then, $\mb{M}_t^i$ is handed to the Representation Disentanglement module which performs useful information disentanglement and outputs $\mb{M}_{t,u}^{i}$. Finally, the motion feature $\mb{M}_{t,u}^{i}$ and the visual feature $\mb{F}_{t}^{i,j}$ and are used to obtain the final pose estimation $\mb{H}_t^i$.}}\label{fig:pipeline}
\vspace{-1.2em}

\end{figure*}

\textbf{Feature difference sequences generation.}\quad We build TDE upon the HRNet-W48 \cite{sun2019deep} network, which includes four convolutional stages to extract feature maps of the input sequence $\{ \mb{F}_{t-1}^{i,J}, \mb{F}_{t}^{i,J}, \mb{F}_{t+1}^{i,J} \}$. The superscript $J = \{1,2,3,4\}$ refers to network stages. Subsequently, we compute the consecutive feature difference sequences $\bm{S}_t^{i,J} = \left \{ \mb{S}^1, \mb{S}^2, \mb{S}^3, \mb{S}^4 \right \}$ over four stages as follows:
\begin{equation}
	\begin{aligned}
		 \mb{S}^j &= \left \{\mb{F}_{t}^{i,j}- \mb{F}_{t-1}^{i,j}, \mb{F}_{t+1}^{i,j} - \mb{F}_{t}^{i,j} \right\}, j = 1, ..., 4.
	\end{aligned}
\end{equation}

\textbf{Intra-stage feature fusion.}\quad Given the feature difference sequences $\bm{S}_t^{i,J}$, several residual blocks \cite{he2016deep} are leveraged to separately aggregate the feature elements within each stage to generate stage-specific motion representations $\bm{D}_t^{i,J} = \left \{ \mb{D}^{1}, \mb{D}^2, \mb{D}^3, \mb{D}^4 \right \}$. This computation can be expressed as:
\begin{equation}
	\begin{aligned}
		 \mb{D}^{j} &= \textbf{Conv}\left [\left(\mb{F}_{t}^{i,j}- \mb{F}_{t-1}^{i,j} \right) \oplus \left(\mb{F}_{t+1}^{i,j} - \mb{F}_{t}^{i,j} \right)\right],
	\end{aligned}
\end{equation}
where $\oplus$ is the concatenation operation and $\mb{Conv}(\cdot)$ is the function of convolutional blocks. In practice, we employ $\left \{ 3, 3, 2, 2 \right\}$ residual blocks with kernel size $3 \times 3$ to aggregate the features at corresponding stages, respectively.

\textbf{Inter-stage feature fusion.}\quad After obtaining the motion features of each stage $\bm{D}_t^{i,J}$, we perform feature integration across stages to obtain the fused motion representation $\mb{M}_t^i$, via the proposed \emph{spatial modulation} and \emph{progressive accumulation}. 
\textbf{(1)} We first employ deformable convolutions (DCN V2 \cite{zhu2019deformable}) to adaptively modulate the spatial-wise responses of each stage feature. Specifically, given $\bm{D}_t^{i,J}$, we independently estimate the kernel sampling offsets $\mathcal{O}$ and modulated scalars $\mathcal{W}$:
\begin{equation}
\begin{aligned}
\left\{\mb{D}^{1}, \cdots, \mb{D}^4 \right\}  &\xrightarrow[\text{blocks}]{\text{residual}} \xrightarrow[\text{convolution}]{\text{regular}} \left\{O^1, \cdots, O^4\right\},\\
\left\{\mb{D}^{1}, \cdots, \mb{D}^4 \right\}  &\xrightarrow[\text{blocks}]{\text{residual}} \xrightarrow[\text{convolution}]{\text{regular}} \left\{W^1, \cdots, W^4\right\}. 
\end{aligned}
\end{equation}
The adaptively learned offsets $\mathcal{O}$ reveal the pixel movement association fields while the modulated scalars $\mathcal{W}$ reflect the magnitude of motion information in each pixel location. Then, we apply deformable convolutions which take the motion features $\bm{D}_t^{i,J}$, the kernel offsets $\mathcal{O}$, and the modulated scalars $\mathcal{W}$ as input, and output the spatially calibrated motion features of each stage $\bar{\bm{D}}_t^{i,J} =\left \{ \bar{\mb{D}}^{1}, \bar{\mb{D}}^2, \bar{\mb{D}}^3, \bar{\mb{D}}^4 \right \}$:
\begin{equation}
\begin{aligned}
\left (\mb{D}^{j}, O^j, W^j \right) \xrightarrow[\text{convolution}]{\text{deformable}} \bar{\mb{D}}^j, j = 1,...,4.
\end{aligned}
\end{equation}
\textbf{(2)} As shown in Fig. \ref{fig:pipeline}, after the spatial modulation, TDE adds the motion feature in the previous stage to the modulated feature, followed by several $3\times3$ convolutions. Such processing is executed progressively until all stages of features are converged to $\mb{M}_t^i$. The above computation is formulated as:
\begin{equation}
\begin{aligned}
\mb{M}_t^i = \textbf{Conv}\left(\cdots\left(\textbf{Conv}\left(\bar{\mb{D}}^1\right) + \cdots + \bar{\mb{D}}^4\right)\right).
\end{aligned}
\end{equation}
By selectively and comprehensively aggregating multi-stage features, our TDE is able to encode informative and fine-grained motion representation $\mb{M}_t^i$.

\subsection{Representation Disentanglement Module} \label{sec. drl}
Directly leveraging the encoded motion feature $\mb{M}_t^i$ for subsequent pose estimation is prone to inevitable task-irrelevant pixel movements  (\emph{e.g.}, background, occlusion). To alleviate this limitation, one can  train a Vision Attention module end-to-end via a heatmap loss to further distill meaningful motion cues. While being straightforward, the learned features in this approach tend to be plain and undistinguished, which results in limited performance improvement (see Table \ref{abl-drl}). After manual examination of the extracted temporal dynamic features for pose estimation, it would be fruitful to investigate whether introducing \emph{supervision to the meaningful information distillation} would facilitate the task.

Ideally, incorporating motion information annotations as feature constraints would simplify this problem. Unfortunately, such movement labels in videos are typically absent in most cases. In light of the above observations, we propose to approach this issue through a mutual information viewpoint. In specific, we explicitly define both useful composition $\mb{M}_{t,u}^{i}$ and noisy composition $\mb{M}_{t,n}^{i}$ of the vanilla motion feature $\mb{M}_{t}^{i}$, in which $\mb{M}_{t,u}^{i}$ is used for subsequent task while $\mb{M}_{t,n}^{i}$ serves as a contrastive landmark. By introducing mutual information supervision to reduce the statistical dependence between $\mb{M}_{t,u}^{i}$ and $\mb{M}_{t,n}^i$, we can grasp discriminative task-relevant motion signals. Coupling these network objectives, the representation disentanglement module (RDM) is designed.

\textbf{Representation factorization.}\quad
Given the vanilla motion representation $\mb{M}_t^i$, we factorize it into useful motion component $\mb{M}_{t,u}^{i}$ and noisy component $\mb{M}_{t,n}^{i}$ by activating corresponding feature channels. Specifically, we first squeeze the global spatial information into a channel descriptor via a global average pooling (GAP) layer. Then, a Multilayer Perceptron (MLP) is used to capture channel-wise interactions followed by a sigmoid function to output an attention mask. This channel-wise attention matrix  is finally exploited to rescale the input feature $\mb{M}_t^i$ to output $\mb{M}_{t,u}^{i}$ and $\mb{M}_{t,n}^{i}$, respectively. The above factorization process is formulated as:
\begin{equation}
\begin{aligned}
\mb{M}_{t,u}^{i} = \sigma\left(\textbf{MLP}_u\left(\textbf{GAP}\left(\mb{M}_t^i\right) \right)\right) \odot \mb{M}_t^i,\\
\mb{M}_{t,n}^{i} = \sigma\left(\textbf{MLP}_n\left(\textbf{GAP}\left(\mb{M}_t^i\right) \right)\right) \odot \mb{M}_t^i.
\end{aligned}
\end{equation}
The symbol $\sigma$ denotes the sigmoid function and $\odot$ refers to the channel-wise multiplication. The network parameters of $\textbf{MLP}_u$ and $\textbf{MLP}_n$ are learned independently. 

\textbf{Heatmap generation.}\quad We integrate the useful motion feature $\mb{M}_{t,u}^{i}$ and the visual feature of the key frame $\mb{F}_{t}^{i,j}$ through several residual blocks to obtain the enhanced representation $ \tilde {\mb{F}}_t^i$. $ \tilde {\mb{F}}_t^i$ is fed to a detection head to yield the estimated pose heatmaps $\mb{H}_t^i$. We implement the detection head using a $3\times 3$ convolutional layer.

\textbf{Mutual information objective.} \quad \label{MI}
Mutual information (MI) measures the amount of information one variable reveals about the other \cite{liu2022temporal, hou2021disentangled}. Formally, the MI between two random variables $\bs{x}_1$ and $\bs{x}_2$ is defined as:
\begin{equation}
\begin{aligned}
	\mathcal{I}(\bs{x}_{1} ; \bs{x}_{2}) = 
	\mathbb{E}_{p(\bs{x}_{1}, \bs{x}_{2})}\left[\log \frac{p(\bs{x}_{1}, \bs{x}_{2})}{p(\bs{x}_{1}) p(\bs{x}_{2})}\right],
\end{aligned}
\end{equation}
where $p(\boldsymbol{x}_{1}, \boldsymbol{x}_{2})$ is the joint probability distribution between $\boldsymbol{x}_{1}$ and $\boldsymbol{x}_{2}$,
while $p(\boldsymbol{x}_{1})$ and $p(\boldsymbol{x}_{2})$ are their marginals.
Within this framework, our main objective for learning effective temporal differences is formulated as: 
\begin{equation}\label{eq8}
	\begin{aligned}
			 \min \text{ } \mathcal{I}\left(\mb{M}_{t,u}^{i}; \mb{M}_{t,n}^{i}  \right) .
	\end{aligned}
\end{equation}
The term $\mathcal{I}\left(\mb{M}_{t,u}^{i}; \mb{M}_{t,n}^{i}  \right)$ rigorously quantifies the amount of information  shared between the useful motion feature $\mb{M}_{t,u}^{i}$ and the noisy feature $\mb{M}_{t,n}^{i}$. Recall that $\mb{M}_{t,u}^{i}$ is used for the subsequent pose estimation task. Intuitively, at the beginning of model training, both meaningful and noisy motion cues will be encoded into $\mb{M}_{t,u}^{i}$ and $\mb{M}_{t,n}^{i}$ simultaneously under the constraint of mutual information minimization. As the training progresses, the features encoded by $\mb{M}_{t,u}^{i}$ and $\mb{M}_{t,n}^{i}$ are gradually systematized into task-relevant motion information and irrelevant motion clues. Under this contrastive setting of $\mb{M}_{t,n}^{i}$, the useful motion feature $\mb{M}_{t,u}^{i}$ would be more discriminative and beneficial for the pose estimation task, as shown in Fig. \ref{fig:motivation} (c).

Furthermore, two regularization functions are introduced to facilitate model optimization. \textbf{(i)} We propose to improve the capacity of the multi-stage temporal difference encoder (TDE) to perceive meaningful information:

\begin{equation}\label{re1}
	\begin{aligned}
			 \max \text{ } \mathcal{I}\left(\mb{M}_{t}^{i}; \mb{M}_{t,u}^{i}  \right) ,
	\end{aligned}
\end{equation}
 where $\mathcal{I}\left(\mb{M}_{t}^{i}; \mb{M}_{t,u}^{i}\right)$ measures the useful motion information compressed in the vanilla motion feature $\mb{M}_{t}^{i}$. Maximizing  this term encourages more abundant and effective motion cues to be encoded in our TDE. 

\textbf{(ii)} We propose to mitigate the information dropping during the feature enhancement of $\tilde {\mb{F}}_t^i$:
\begin{equation}
\begin{aligned}
\label{eq.re.}
\min \left[ 
\mathcal{I}\left(\mb{M}_{t,u}^{i};{y}^{i}_{t}\mid\tilde {\mb{F}}_t^i  \right)
+ 
\mathcal{I}\left(\mb{F}_{t}^{i,j};{y}^{i}_{t}\mid\tilde{\mb{F}}_t^i  \right)
\right],
\end{aligned} 
\end{equation}

where ${y}^{i}_{t}$ refers to the label, and terms $\mathcal{I}\left(\mb{M}_{t,u}^{i};{y}^{i}_{t} \mid    \tilde {\mb{F}}_t^i  \right)$ and $\mathcal{I}\left(\mb{F}_{t}^{i,j};{y}^{i}_{t}\mid\tilde{\mb{F}}_t^i  \right)$ quantify the vanishing task-relevant information in $\mb{M}_{t,u}^{i}$ and $\mb{F}_{t}^{i,j}$ respectively when aggregating them into the enhanced feature $\tilde {\mb{F}}_t^i$. Minimizing these two regularization terms fosters the nondestructive propagation of information. Given the notorious difficulty of conditional MI calculations  \cite{hjelm2018learning, tian2021farewell}, we adopt a simplified version as done in \cite{liu2022temporal, zhao2021learning}. We approximate the first term $\mathcal{I}\left(\mb{M}_{t,u}^{i};{y}^{i}_{t} \mid    \tilde {\mb{F}}_t^i  \right)$ as:

\begin{equation}
\begin{aligned} 
\mathcal{I}\left(\mb{M}_{t,u}^{i};{y}^{i}_{t}  \mid \tilde{\mb{F}}_t^i  \right) \rightarrow \mathcal{I}\left({\mb{M}_{t,u}^{i};y}^{i}_{t} \right) - \mathcal{I}\left(\mb{M}_{t,u}^{i} ; \tilde{\mb{F}}_t^i\right).
\end{aligned} 
\end{equation}
Similarly, the second term $\mathcal{I}\left(\mb{F}_{t}^{i,j};{y}^{i}_{t}\mid\tilde{\mb{F}}_t^i  \right)$ in Eq. \ref{eq.re.} can be simplified as follows:
\begin{equation}
\begin{aligned} 
\mathcal{I}\left(\mb{F}_{t}^{i,j};{y}^{i}_{t}\mid\tilde{\mb{F}}_t^i  \right)\rightarrow 
\mathcal{I}\left(\mb{F}_{t}^{i,j}; y^{i}_{t} \right) - 
\mathcal{I}\left(\mb{F}_{t}^{i,j} ; \tilde{\mb{F}}_t^i\right).
\end{aligned} 
\end{equation}

Ultimately, the proposed primary objective (Eq. \ref{eq8}) and the regularization terms (Eq. \ref{re1} and Eq. \ref{eq.re.}) are simultaneously optimized to provide feature supervision:
\begin{equation}
\begin{aligned} 
\mathcal{L}_{\text{MI}} &= \mathcal{I}\left(\mb{M}_{t,u}^{i}; \mb{M}_{t,n}^{i}  \right) -  \mathcal{I}\left(\mb{M}_{t}^{i}; \mb{M}_{t,u}^{i}  \right) \\ &+ \mathcal{I}\left(\mb{M}_{t,u}^{i};{y}^{i}_{t}\mid\tilde {\mb{F}}_t^i  \right)
+ 
\mathcal{I}\left(\mb{F}_{t}^{i,j};{y}^{i}_{t}\mid\tilde{\mb{F}}_t^i  \right).
\end{aligned} 
\end{equation}
We adopt Variational Self-Distillation (VSD) \cite{tian2021farewell} to compute the MI for each term.

\subsection{Loss Functions}
Overall, our loss function consists of two portions. (1) We employ the standard pose heatmap loss $\mathcal{L}_\text{H}$ to supervise the final pose estimation:
\begin{equation}
	\begin{aligned}
		\mathcal{L}_\text{H} = \left\|\mb{H}_t^i - \mb{G}_t^i \right\| _{2}^{2},
	\end{aligned}
\end{equation}
where $\mb{H}_t^i$ and $\mb{G}_t^i$ denote the predicted and ground truth pose heatmaps, respectively. (2) We also leverage the proposed mutual information objective $\mathcal{L}_\text{MI}$ to supervise the learning of motion features. Our total loss is given by:
\begin{equation}\label{eq.final}
	\begin{aligned}
		\mathcal{L}_\text{total} = \mathcal{L}_\text{H} + \alpha \mathcal{L}_\text{MI},
	\end{aligned}
\end{equation}
where $\alpha$ is a hyper-parameter to balance the ratio of different loss terms.
\section{Experiments}

\renewcommand\arraystretch{1.1}
\begin{table}
  \resizebox{0.48\textwidth}{!}{
  \begin{tabular}{l|ccccccc|c}
    \hline
      Method                            &Head   &Shoulder &Elbow       &Wrist   &Hip    &Knee   &Ankle   &{\bf Mean}\cr
      \hline
      PoseTracker \cite{girdhar2018detect}   &$67.5$ &$70.2$   &$62.0$      &$51.7$  &$60.7$ &$58.7$ &$49.8$  &{$60.6$}\cr
     PoseFlow \cite{xiu2018pose}         &$66.7$ & $73.3$  &$68.3$      &$61.1$  &$67.5$ &$67.0$ &$61.3$  &{$ 66.5$}\cr
JointFlow \cite{doering2018joint}        & -     & -       &-           &-       &-      &-      &-       &{ $ 69.3$}\cr
   FastPose \cite{zhang2019fastpose}   	&$80.0$ &$80.3$   &$69.5$      &$59.1$  &$71.4$ &$67.5$ &$59.4$  &{$ 70.3$}\cr
   TML++ \cite{hwang2019pose}    	 		&-       &-     &-      &-      &-      &-       &-    &{$ 71.5$}\cr
Simple (R-50) \cite{xiao2018simple}    &$79.1$ &$80.5$   &$75.5$      &$66.0$  &$70.8$ &$70.0$ &$61.7$  &{$72.4$}\cr
Simple (R-152) \cite{xiao2018simple}    &$81.7$ &$83.4$   &$80.0$      &$72.4$  &$75.3$ &$74.8$ &$67.1$  &{$ 76.7$}\cr
  STEmbedding \cite{jin2019multi}        &$83.8$ &$81.6$   &$77.1$      &$70.0$  &$77.4$ &$74.5$ &$70.8$  &{$ 77.0$}\cr
        HRNet \cite{sun2019deep}         &$82.1$ &$83.6$   &$80.4$      &$73.3$  &$75.5$ &$75.3$ &$68.5$  &{$ 77.3$}\cr
         MDPN \cite{guo2018multi}        &$85.2$ &$88.5$   &$83.9$      &$77.5$  & $79.0$&$77.0$ &$71.4$  &{$ 80.7$}\cr
   CorrTrack \cite{rafi2020self}   &$86.1$ &$87.0$   &$83.4$      &$76.4$  & $77.3$&$79.2$ &$73.3$  &{$ 80.8$}\cr 
   Dynamic-GNN \cite{yang2021learning} 	 &$88.4$ &$88.4$   &$82.0$      &$ 74.5$ &$79.1$ &$78.3$ &$73.1$  &{ $81.1$}\cr
   PoseWarper \cite{bertasius2019learning} &$81.4$ &$88.3$   &$83.9$      &$ 78.0$ &$82.4$ &$80.5$ &$73.6$  &{ $ 81.2$}\cr
   DCPose \cite{liu2021deep}  &$ 88.0$  &$ 88.7$     &$ 84.1$   &$78.4$&$ 83.0$        &$ 81.4$&$ 74.2$ &$ 82.8$\cr
   DetTrack \cite{wang2020combining}  &$89.4$       &$89.7$     &$85.5$ &$79.5$ &$82.4$      &$80.8$       &$76.4$   &$83.8$\cr
    FAMI-Pose \cite{liu2022temporal}	&$ 89.6$  &$ 90.1$ &$ 86.3$ &$80.0$ &$ 84.6$ &$83.4$ &$ 77.0$ &$ 84.8$\cr
    \hline 
    \rowcolor{gray!28}  \bf TDMI (Ours)	&$ 90.0$  &$\bf 91.1$ &$ 87.1$ &$ 81.4$ &$\bf 85.2$ &$ \bf 84.5$ &$ 78.5$ &$ 85.7$\cr 
     \rowcolor{gray!28} \bf TDMI-ST (Ours)	&$\bf 90.6$  &$ 91.0$ &$\bf 87.2$ &$\bf 81.5$ &$\bf 85.2$ &$\bf 84.5$ &$\bf 78.7$ &$\bf 85.9$\cr 
    \hline
    \end{tabular}}
    \vspace{-0.5em}
    \caption{Quantitative results on the \textbf{PoseTrack2017} validation set.} \label{17val}\vspace{-0.2em}
\end{table}

\renewcommand\arraystretch{1.2}
\begin{table}
   \resizebox{0.48\textwidth}{!}{
   \begin{tabular}{l|ccccccc|c}
     \hline
      Method                            &Head &Shoulder &Elbow  &Wrist &Hip &Knee &Ankle &{\bf Mean}\cr
     \hline
  STAF \cite{raaj2019efficient}    	  	&-       &-     &-      &$64.7$ &-      &-       &$62.0$   &{$70.4$}\cr
 AlphaPose \cite{fang2017rmpe}           &$63.9$  &$78.7$&$77.4$ &$71.0$ &$73.7$ &$73.0$    &$69.7$     &{$71.9$}\cr
  TML++ \cite{hwang2019pose}    	 		&-       &-     &-      &-      &-      &-       &-    &{$ 74.6$}\cr
 MDPN \cite{guo2018multi}                &$75.4$ &$81.2$ &$79.0$ &$74.1$ &$72.4$ &$73.0$  &$69.9$   &{$75.0$}\cr
 PGPT \cite{bao2020pose}    	 		&-       &-     &-      &$72.3$ &-      &-       &$72.2$   &{$76.8$}\cr
 Dynamic-GNN \cite{yang2021learning} 	&$80.6$ &$84.5$   &$80.6$  &$ 74.4$ &$75.0$ &$76.7$ &$71.8$  &{ $77.9$}\cr
 PoseWarper \cite{bertasius2019learning} &$79.9$&$86.3$&$82.4$&$77.5$&$79.8$&$78.8$&$73.2$  &{ $79.7$}\cr
 PT-CPN++ \cite{yu2018multi} &$82.4$ &$88.8$ &$86.2$ &$79.4$ &$72.0$ &$80.6$ &$76.2$  &$80.9$\cr
 DCPose \cite{liu2021deep}&$ 84.0$ &$ 86.6$&$ 82.7$&$ 78.0$&$ 80.4$&$ 79.3$&$ 73.8$&$ 80.9$\cr 
 DetTrack \cite{wang2020combining}  &$84.9$ &$87.4$ &$84.8$ &$79.2$ &$77.6$      &$79.7$       &$75.3$   &$81.5$ \cr 
 FAMI-Pose \cite{liu2022temporal} &$ 85.5$&$ 87.7$&$ 84.2$&$ 79.2$&$ 81.4$&$81.1$&$ 74.9$&$ 82.2$\cr
	 \hline
    \rowcolor{gray!28}  \bf TDMI (Ours)&$ 86.2$&$ 88.7$&$\bf 85.4$&$\bf 80.6$&$\bf 82.4$&$\bf 82.1$&$ 77.5$&$ 83.5$\cr
    \rowcolor{gray!28}  \bf TDMI-ST (Ours)&$\bf 86.7$&$\bf 88.9$&$\bf 85.4$&$\bf 80.6$&$\bf 82.4$&$\bf 82.1$&$\bf 77.6$&$\bf 83.6$\cr
     \hline
     \end{tabular}}
     \vspace{-0.5em}
     \caption{Quantitative results on the \textbf{PoseTrack2018} validation set.} \label{18val}\vspace{-0.2em}
   \end{table}

\renewcommand\arraystretch{1.2}
\begin{table}
   \resizebox{0.48\textwidth}{!}{
   \begin{tabular}{l|ccccccc|c}
     \hline
      Method                            &Head &Shoulder &Elbow  &Wrist &Hip &Knee &Ankle &{\bf Mean}\cr
     \hline
   Tracktor++  w. poses \cite{bergmann2019tracking, doering2022posetrack21}   	  	&-       &-     &-      &- &-      &-       &-   &{$71.4$}\cr
  CorrTrack \cite{rafi2020self, doering2022posetrack21}    	  	&-       &-     &-      &- &-      &-       &-   &{$72.3$}\cr
  CorrTrack w. ReID \cite{rafi2020self, doering2022posetrack21}    	  	&-       &-     &-      &- &-      &-       &-   &{$72.7$}\cr
  Tracktor++ w. corr. \cite{bergmann2019tracking, doering2022posetrack21}    	  	&-       &-     &-      &- &-      &-       &-   &{$73.6$}\cr
 DCPose \cite{liu2021deep}&$ 83.2$&$ 84.7$&$ 82.3$&$ 78.1$&$ 80.3$&$ 79.2$&$ 73.5$&$ 80.5$\cr 
 FAMI-Pose \cite{liu2022temporal} &$ 83.3$&$ 85.4$&$ 82.9$&$ 78.6$&$ 81.3$&$80.5$&$ 75.3$&$ 81.2$\cr
	 \hline
     \rowcolor{gray!28}  \bf TDMI (Ours)&$ 85.8$&$\bf 87.5$&$\bf 85.1$&$ 81.2$&$ 83.5$&$82.4$&$ 77.9$&$ 83.5$\cr
     \rowcolor{gray!28} \bf TDMI-ST (Ours)&$\bf 86.8$&$ 87.4$&$\bf 85.1$&$\bf 81.4$&$\bf 83.8$&$\bf 82.7$&$\bf 78.0$&$\bf 83.8$\cr
     \hline
     \end{tabular}}
     \vspace{-0.5em}
     \caption{Quantitative results on the \textbf{PoseTrack21} dataset.} \label{21val}
     \vspace{-0.5em}
   \end{table}

\subsection{Experimental Settings}
\textbf{Datasets.}\quad
PoseTrack is a large-scale benchmark dataset for video-based human pose estimation.
Specifically, \textbf{PoseTrack2017} contains $250$ video sequences for training and $50$ video sequences for validation (following the official protocol), with a total of $80,144$ pose annotations. \textbf{PoseTrack2018} greatly increases the number of videos and includes $593$ for training and $170$ for validation (with $153,615$ pose annotations). Both datasets are annotated with $15$ keypoints, with an additional flag for joint visibility. \textbf{PoseTrack21} further extends pose annotations of the PoseTrack2018 dataset, especially for particular small persons and persons in crowds, including $177,164$ pose annotations. The flag of joint visibility is re-defined in PoseTrack21 so that occlusion information can be utilized. \textbf{HiEve} \cite{lin2020human} is a very challenging benchmark dataset for human-centric video analysis in various realistic crowded and complex events (\emph{e.g.}, earthquake, getting-off train, and bike collision), containing $32$ video clips where $19$ for training and $13$ for testing. This dataset possesses  a substantially larger data scale and includes the currently \emph{largest} number of pose annotations ($1,099,357$). 

\renewcommand\arraystretch{1.4}
\begin{table}
   \resizebox{0.48\textwidth}{!}{
   \begin{tabular}{l|cccc|cccc}
     \hline
      Method &\bf w\_AP\text{@}avg &w\_AP\text{@}50 &w\_AP\text{@}75  &w\_AP\text{@}90 &\bf AP\text{@}avg &AP\text{@}50 &AP\text{@}75  &AP\text{@}90 \cr
     \hline
  RSN18 \cite{cai2020learning}    	 &$48.3$  &$52.5$&$48.9$ &$43.6$ &$52.4$ &$56.2$    &$53.2$     &$47.8$	   \cr
  DHRN FT \cite{sun2019deep}         &$51.6$  &$59.6$&$50.1$ &$45.3$ &$55.3$ &$63.0$ 	&$53.9$ 	&$49.1$    \cr
 ADAM+PRM \cite{lin2020human}        &$54.0$  &$63.0$&$52.2$ &$46.7$ &$57.7$ &$66.5$ 	&$55.9$ 	&$50.6$    \cr
 DH\_IBA \cite{yuan2020combined}      &$55.2$  &$66.7$&$52.5$ &$46.3$ &$59.3$     &$70.5$ 	&$56.8$ 	&$50.5$    \cr
 TryNet \cite{liu2021deep}			 &$55.2$  &$68.7$&$51.7$ &$45.3$ &$59.3$ &$72.6$ 	&$56.1$ 	&$49.3$    \cr 
 ccc$\dag$ \cite{chang2020towards}		 &$56.3$  &$\bf 69.6$&$53.0$ &$46.4$ &$60.0$ &$\bf 73.1$ 	&$56.9$ 	&$49.9$    \cr 
	 \hline
    \rowcolor{gray!28}  \bf TDMI-ST (Ours) &$\bf 58.5$ &$69.0$&$\bf 56.2$&$\bf 50.4$&$\bf 62.1$&$ 72.4$&$\bf 60.0$&$\bf 54.0$\cr
     \hline
     \end{tabular}}
     \vspace{-0.5em}
     \caption{Quantitative results from the \textbf{HiEve} \emph{test leaderboard}. $\dag$ means using extra data. $\text{AP@}\beta$ denotes the AP value calculated under given distance threshold $\beta$, and the prefix ``w" indicates that the value is calculated with frame weight. The official leaderboard ranks different approaches leveraging the metric of \textbf{w\_AP\text{@}avg}.} \label{hieve}
     \vspace{-1em}
   \end{table}

\textbf{Evaluation metric.}\quad
We employ the standard pose estimation metric namely average precision (\textbf{AP}) to evaluate our model. We first compute the AP for each joint  and then obtain the final performance (\textbf{mAP}) by averaging all joints.

\textbf{Implementation details.}
Our TDMI framework is implemented with PyTorch. For visual feature extraction, we leverage the HRNet-W48 \cite{sun2019deep} model that is pre-trained on the COCO dataset. We incorporate several data augmentation strategies as adopted in \cite{sun2019deep, bertasius2019learning, xiao2018simple}, including random rotation $[-45^{\circ}, 45^{\circ}]$, random scale $[0.65, 1.35]$,  truncation (half body), and flipping. The temporal span $\delta$ is set to $2$, \emph{i.e.}, $2$ previous and $2$ future frames. We use the Adam optimizer with an initial learning rate of $1e-3$ (decays to $1e-4$, $1e-5$, and $1e-6$ at the $6^{th}$, $12^{th}$, and $18^{th}$ epochs, respectively). We train the model using $2$ TITAN RTX GPUs. The training process is terminated within $20$ epochs. To weight difference losses in Eq. \ref{eq.final}, we empirically found that $\alpha =1$ is the most effective. During the feature enhancement of $ \tilde {\mb{F}}_t^i$, the final stage feature $\mb{F}_{t}^{i,4}$ is adopted as the visual feature.
\begin{figure*}
\begin{center}
\includegraphics[width=0.96\linewidth]{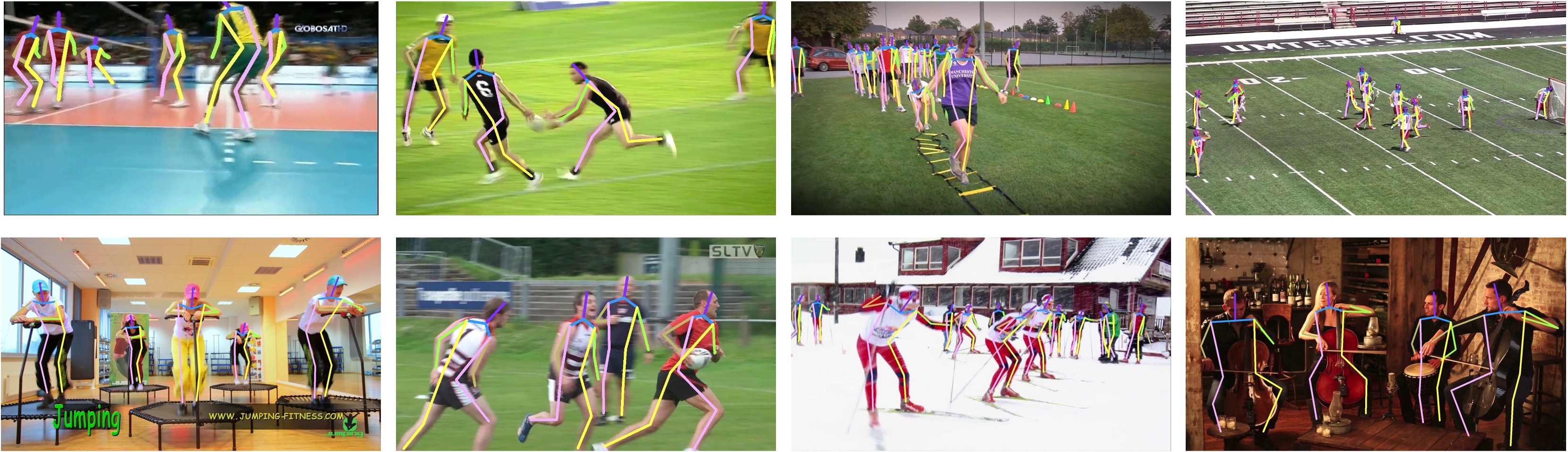}
\end{center}
\vspace{-1.2em}
\caption{Visual results of our TDMI on benchmark datasets. Challenging scenes such as fast motion or pose occlusion are involved.}
\vspace{-1.5em}
\label{fig:results}
\end{figure*}

\subsection{Comparison with State-of-the-art Approaches}
\textbf{Results on the PoseTrack2017 dataset.}\quad
We first evaluate our method on the PoseTrack2017 dataset. A total of $18$ methods are compared and their performances on the validation set are reported in Table \ref{17val}. Our TDMI model consistently outperforms existing state-of-the-art methods, reaching an mAP of $85.7$. To fully leverage  temporal clues, we extend the TDMI to \textbf{TDMI-ST} where we replace the keyframe visual feature with the spatiotemporal feature of the input sequence to yield the final representation $\tilde {\mb{F}}_t^i$. The proposed TDMI-ST further pushes forward the performance boundary and achieves an mAP of $85.9$. Remarkably, our TDMI-ST improves mAP by $8.6$ points over the adopted backbone network HRNet-W48 \cite{sun2019deep}. Compared to the previous best-performed method FAMI-Pose \cite{liu2022temporal}, our TDMI-ST  also delivers a $1.1$ mAP gain. The performance boost for challenging joints (\emph{i.e.}, wrist, ankle) is also encouraging: we obtain an mAP of $81.5$ ($\uparrow 1.5$) for wrists and an mAP of $78.7$ ($\uparrow 1.7$) for ankles. Such consistent and significant performance improvements suggest the importance of explicitly embracing meaningful motion information. Moreover, we display the visualized results for scenarios with complex spatio-temporal interactions (\emph{e.g.}, occlusion, blur) in  Fig. \ref{fig:results}, which attest to the robustness of the proposed method. 
\textbf{Results on the PoseTrack2018 dataset.}\quad
We further benchmark our model on the PoseTrack2018 dataset. Empirical comparisons on the validation set are tabulated in Table \ref{18val}. As presented in this table, our basic model, TDMI, surpasses all other approaches and achieves an mAP of $83.5$.  Our TDMI-ST model, on the other hand, attains the new state-of-the-art results over all joints and obtains the final performance of $83.6$ mAP, with an mAP of $85.4$, $80.6$, and $77.6$ for the elbow, wrist, and ankle, respectively.

\textbf{Results on the PoseTrack21 dataset.}\quad
Table \ref{21val} reports the results of our method as well as other state-of-the-art methods on the PoseTrack21 dataset. Quantitive results of the first four baselines \cite{bergmann2019tracking, rafi2020self, doering2022posetrack21} are officially provided by the dataset \cite{doering2022posetrack21}. We further reproduce several previous impressive approaches (\emph{i.e.},  DCPose \cite{liu2021deep} and FAMI-Pose \cite{liu2022temporal}) according to their released implementations on GitHub, and evaluate their performances in this dataset. From Table \ref{21val}, we observe that FAMI-Pose \cite{liu2022temporal} achieves favorable pose estimation results with an mAP of $81.2$. In contrast, our TDMI and TDMI-ST are able to obtain $83.5$ mAP and $83.8$ mAP, respectively. Another observation is that PoseTrack21 mainly increases the pose annotations of  small persons and persons in crowds for PoseTrack2018. Interestingly, the proposed TDMI-ST attains a better performance on PoseTrack21 with respect to PoseTrack2018 ($\uparrow 0.2$ mAP), which might be an evidence to show the effectiveness and robustness of our method especially for challenging scenes.

\textbf{Results on the HiEve dataset.}\quad
Furthermore, we evaluate our model on the \emph{largest} HiEve benchmark dataset. The detailed results of the test set are tabulated in Table \ref{hieve}. We upload the prediction results of our model to the HiEve test  server\footnote{\url{http://humaninevents.org/oltp.html?title=3}} to obtain results. Our TDMI-ST achieves top scoring weight-average AP (w\_AP\text{@}avg) of $58.5$ on the HiEve leaderboard. Significantly, compared to ccc$\dag$ \cite{chang2020towards} that uses extra data to train the model, we still achieve performance improvements by $2.2$ w\_AP and $2.1$ AP, respectively.  

\begin{table*}
    \centering 
	\begin{minipage}[t]{0.21\linewidth}
	\vspace{0pt}
    \centering 
	\resizebox{1\textwidth}{!}{
     \begin{tabular}{c|cc|cc|c}
    \hline
     Method &TDE &RDM &Mean\cr
    \hline
    HRNet \cite{sun2019deep} & &   & $77.3$\cr
    Op-Flow & &   & $84.0$\cr
    (a) &\checkmark &  & $84.5$\cr
    (b) &\checkmark &\checkmark  & $85.7$\cr
    \hline
    \end{tabular}}
    \vspace{-0.5em}
   \caption{\footnotesize{Ablation of different components in \textbf{TDMI}}.} 
   \label{abl-com} 
  \end{minipage}
  \begin{minipage}[t]{0.4\linewidth} 
  	\vspace{0pt}
  \centering
	\resizebox{1\textwidth}{!}{
  	\begin{tabular}{c|ccc|c}
    \hline
     Method &Multi-stage  &Spatial modulation &Progressive fusion &Mean\cr
    \hline
    (a) & & &  & $84.4$\cr
    (b) &\checkmark & &  & $84.6$\cr
   	(c) &\checkmark & &\checkmark  & $84.9$\cr
   	(d) &\checkmark &\checkmark &\checkmark & $85.7$\cr
    \hline
    \end{tabular}}
    \vspace{-0.5em}
    \caption{\footnotesize{Ablation of various designs in \textbf{TDE}.}}
    \label{abl-TDE}
  \end{minipage}
 \begin{minipage}[t]{0.315\linewidth} 
 	\vspace{0pt}
	 \centering
 	\resizebox{1\textwidth}{!}{
 	\begin{tabular}{c|cc|c}
    \hline
     Method &Factorization  &MI objective &Mean\cr
    \hline
    TDMI, w/o RDM & & & $84.5$\cr
    (a) &\checkmark &   & $84.7$\cr
   	(b) &\checkmark &\checkmark  & $85.7$\cr
   	(c)$\ddag$ &\checkmark &\checkmark  & $83.8$\cr
    \hline
    \end{tabular}}
    \vspace{-0.5em}
    \caption{\footnotesize{Ablation of various designs in \textbf{RDM}. ``w/o X" refers to removing X module}.} 
    \label{abl-drl}
 \end{minipage}
\vspace{-0.5em}
\end{table*}

\begin{figure}
\begin{center}
\includegraphics[width=0.98\linewidth]{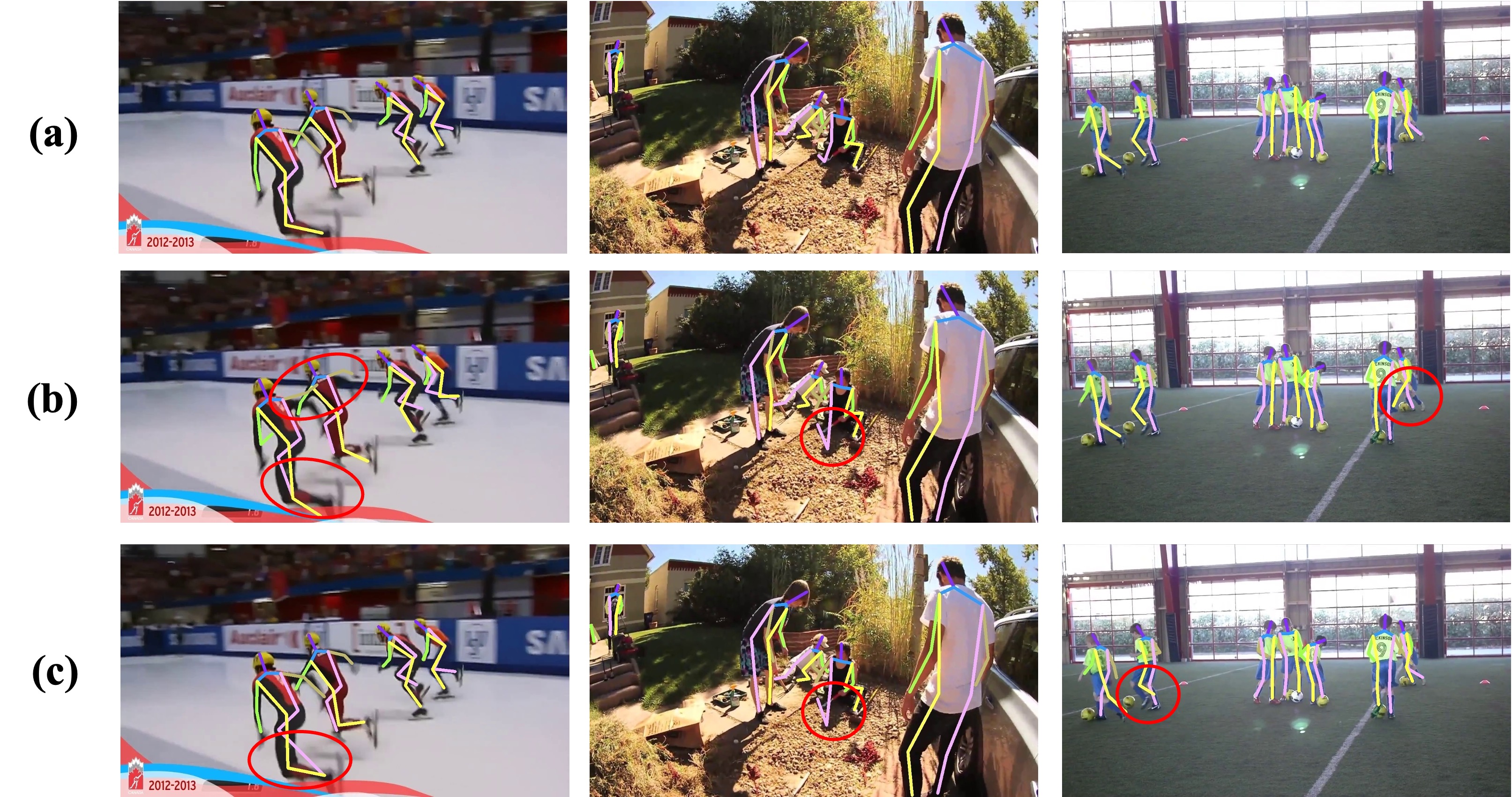}
\end{center}
\vspace{-1.5em}
\caption{Qualitative comparisons of the prediction results of our TDMI (a), HRNet-W48 (b),  and FAMI-Pose (c) on the challenging cases from PoseTrack dataset. Inaccurate detections are highlighted by the red circles.}
\label{fig:qualitative}
\vspace{-1.7em}
\end{figure}

\textbf{Comparison of visual results.}\quad
In addition to the quantitative analysis, we also qualitatively examine the capability of our approach in handling complex situations such as occlusion and blur. We depict in Fig. \ref{fig:qualitative} the side-by-side comparisons of a) our TDMI against state-of-the-art methods b) HRNet \cite{sun2019deep} and c) FAMI-Pose \cite{liu2022temporal}. It is observed that our approach consistently provides more accurate and robust pose detection for various challenging scenarios. HRNet is designed for static images and does not incorporate temporal dynamics, resulting in suboptimal results in degraded frames. In contrast, FAMI-Pose performs implicit  motion compensation yet lacks effective information distillation. Through the principled design of TDE and RDM for temporal difference modeling and useful information disentanglement, our TDMI is more adept at handling challenging  scenes such as occlusion and blur.

\subsection{Ablation Study}
We perform ablation studies focused on examining the contribution of each component in our TDMI  framework, including the multi-stage Temporal Difference Encoder (TDE) and the Representation Disentanglement module (RDM). We also investigate the effectiveness of various micro designs within each component. All experiments are conducted on the PoseTrack2017 validation set.

\textbf{Study on components of TDMI.}\quad
We empirically evaluate the efficacy of each component in the proposed TDMI and report quantitative results in Table \ref{abl-com}. Op-Flow is a baseline where we employ optical flows predicted by the RAFT \cite{teed2020raft} as motion representations. \textbf{(a)} For the first setting, we introduce TDE to the HRNet-W48 \cite{sun2019deep} baseline for capturing motion contexts. Remarkably, the motion clues encoded by the TDE already improve over the baseline by a large margin of $7.2$ mAP. This demonstrates the effectiveness of our TDE in introducing motion information to facilitate video-based human pose estimation. On the other hand, our TDE also delivers a performance gain of $0.5$ mAP over the baseline Op-Flow. This highlights the great potential of temporal differences in representing motions as compared to complex optical flow. \textbf{(b)} For the next setting, we further incorporate the RDM to discover meaningful motions. The results in mAP increase to $85.7$ by $1.2$. This significant performance improvement on top of the well-established TDE corroborates the importance of excavating meaningful temporal dynamics in guiding accurate pose estimation.

\textbf{Study on multi-stage Temporal Difference Encoder.}\quad
We modify the TDE with various designs to investigate their influences on the final performance. As reported in Table \ref{abl-TDE}, four experiments are conducted: \textbf{(a)} fusing only the features of the final stage $\mb{S}^4$, \textbf{(b)} directly aggregating multi-stage features via simple concatenation and convolution, \textbf{(c)} progressively integrating multi-stage features, and \textbf{(d)} our complete TDE with spatial modulation and progressive fusion. From this table, we observe that multi-stage feature fusion indeed outperforms using only single-stage features, yet the simple fusion scheme \textbf{(b)} yields a slight performance improvement ($\uparrow 0.2$ mAP). By progressively aggregating multi-stage features \textbf{(c)}, detailed information at each stage is preserved which provides an mAP gain of $0.5$ points. Our complete TDE \textbf{(d)} further incorporates a spatial modulation mechanism to adaptively select important information of each stage, achieving the best performance.

\textbf{Study on Representation Disentanglement module.}\quad
In addition, we examine the effects of RDM under different settings and present the results in Table \ref{abl-drl}. We first conduct a simple baseline \textbf{(a)} in which the MI objective is removed. This method partially distills the useful motion information and marginally improves the mAP by $0.2$ ($84.5 \rightarrow 84.7$). We then incorporate the MI objective, which corresponds to our full RDM \textbf{(b)}. The noticeable performance enhancement of $1.0$ mAP provides empirical evidence that our proposed MI objective is effective as additional supervision to facilitate the learning of discriminative task-relevant motion clues. We also attempt to perform the feature factorization by splitting channels \textbf{(c)$\ddag$} and the mAP drops $1.9$, which suggests that the channel splitting scheme might be inapplicable to our TDMI framework.

\section{Conclusion and Future Works}
In this paper, we investigate the video-based human pose estimation task from the perspective of effectively exploiting dynamic contexts through temporal difference learning and useful information disentanglement. We present a multi-stage Temporal Difference Encoder (TDE) to capture motion clues conditioned on explicit feature difference representation. Theoretically, we further build a Representation Disentanglement module (RDM) on top of mutual information to grasp task-relevant information. Extensive experiments demonstrate that the proposed method outperforms state-of-the-art approaches on four benchmark datasets, including PoseTrack2017, PoseTrack2018, PoseTrack21, and HiEve. Future works include applications to other video-related tasks such as 3D human pose estimation and action recognition. The temporal difference features can also be integrated into existing pose-tracking pipelines to assess the similarity of human  motions for data association.

\section{Acknowledgements}
This work is supported in part by the National Natural Science Foundation of China under grant No. 62203184. This work is also supported in part by the MSIT, Korea, under the ITRC program (IITP-2022-2020-0-01789) (50\%) and the High-Potential Individuals Global Training Program (RS-2022-00155054) (50\%) supervised by the IITP.



{\small
\bibliographystyle{ieee_fullname}
\bibliography{cvpr}
}

\end{document}